\documentclass{article}
\usepackage{spconf,amsmath,graphicx}
\usepackage{xcolor}
\usepackage[english]{babel}
\usepackage{bm}

\title{Generative linguistic representation for spoken language identification}
\name{Peng Shen, Xuguang Lu, Hisashi Kawai}
\address{National Institute of Information and Communications Technology (NICT), Japan}
\begin{document}
%
\maketitle
\begin{abstract}
  Effective extraction and application of linguistic features are central to the enhancement of spoken Language IDentification (LID) performance.
  With the success of recent large models, such as GPT and Whisper, the potential to leverage such pre-trained models for extracting linguistic features for LID tasks has become a promising area of research. In this paper, we explore the utilization of the decoder-based network from the Whisper model to extract linguistic features through its generative mechanism for improving the classification accuracy in LID tasks.
  We devised two strategies - one based on the language embedding method and the other focusing on direct optimization of LID outputs while simultaneously enhancing the speech recognition tasks.
  We conducted experiments on the large-scale multilingual datasets MLS, VoxLingua107, and CommonVoice to test our approach.
  The experimental results demonstrated the effectiveness of the proposed method on both in-domain and out-of-domain datasets for LID tasks.

\end{abstract}
\section{Introduction}
\label{sec:intro}
Language IDentification(LID) in spoken language is an essential component in multilingual speech processing systems. The emergence of deep learning algorithms has accelerated advancements in LID technologies significantly \cite{HaizhouLi2013,DehakIEEE2015, LOPEZMORENO201646, shen2016, LozanoDiez2015, Fernando2017, GengWang2016}. For LID, it's not only the acoustic features like phonotactics information that matter but also linguistic features such as contextual information play a vital role in language determination \cite{HaizhouLi2013, shen2022transducerbased}. The precision of LID has been substantially improved by recent deep neural network-based LID technologies. However, these techniques tend to overfit in real-world applications due to the cross-domain problem, i.e., differences in recording conditions and speaking styles between the training and testing datasets.

Previous studies aimed at enhancing the performance and combating the overfitting issue in LID systems have proposed the use of phonetic and contextual information for system development \cite{lizheng2021additivePhoneme, Ren2019TwostageTF,jicheng2021E2EmultiTask, LiBo2018multiDialectASR}. Most of these studies leverage speech recognition models to bolster the LID system. Ren et al. proposed a two-step training process, which first trains an acoustic model with a connectionist temporal classification (CTC), then a recurrent neural network classifies the language category using the intermediate features derived from the acoustic model as inputs \cite{Ren2019TwostageTF}.
Multi-task training methods have also been investigated, which enhance performance and bolster model robustness. This method utilizes the shared underlying feature extraction network and jointly trains objective functions for speech/phoneme recognition and language recognition \cite{lizheng2021additivePhoneme, jicheng2021E2EmultiTask, LiBo2018multiDialectASR}.
Consideration has also been given to self-supervised phonotactic representations that use context information \cite{ramesh21_interspeech, Fanzhiyun2021wav2vec4LID}.
In addition to the usage of the implicit linguistic expression method, an RNN transducer-based approach uses explicit linguistic expression derived from the prediction of the RNN-T to enhance LID tasks \cite{shen2022transducerbased}.
All these methods illustrate the importance of linguistic features for LID tasks.

Recently, the success of ChatGPT and other large, generative language models has attracted widespread attention. The decoder-based GPT model has demonstrated impressive capabilities not only in text generation but also in image analysis \cite{gpt3,gpt4}. Particularly notable are the substantial improvements in logical analysis skills with increased parameter size. Previous works on speech-based classification tasks primarily utilized models based on transformer encoders, such as HuBERT \cite{huBERT} and wav2vec 2.0 \cite{wav2vec2}. However, the utilization of such a decoder-based generative network for LID tasks remains to be investigated.

In this paper, we investigate the use of the decoder network of the OpenAI Whisper model to improve LID performance. The Whisper model is a transformer-based, multilingual Automatic Speech Recognition (ASR) model that was trained on 680,000 hours of data across 96 languages. Thus, it potentially serves as a pre-trained model for various downstream tasks. We aim to explore and classify how to use the decoder network to extract linguistic representation for LID tasks. To achieve this, we evaluate it using a widely adopted language embedding method and a fine-tuning framework. Finally, to mitigate the 'forgetting problem' during the fine-tuning process, we employ an ASR-enhanced learning method to further improve the generation of linguistic expressions. To the best of our knowledge, no previous work has focused on using decoder-based linguistic representation for LID tasks.
The contributions of this work can be summarized as follows:
\begin{itemize}
  \item We first analyzed the limitations of the Whisper model in handling LID tasks, and proposed two strategies to further leverage its linguistic features to improve LID performance.
  \item To solve the optimization issue of the decoder for LID, we designed a variety of experiments to understand the influence of the language conditional setting. To address the issue of knowledge forgetting during the fine-tuning process, we proposed using an ASR-enhanced learning method.
  \item By comparing the performance of our proposed method on base and large-v2 models, we found that the high-performance large model plays a critical role in extracting linguistic information to enhance LID.
\end{itemize}


\section{Generative linguistic representation}

\subsection{The shortcomings of the Whisper model for LID}
Our investigation is based on the OpenAI Whisper model, which has been scaled to 680,000 hours of multilingual and multitask supervision \cite{whisper2022}. Given the extensive training data, the Whisper model demonstrates superior generalization capabilities, performing well across various datasets.
Although the Whisper model can identify languages, its accuracy is only 80.3\%. Due to this limitation, to obtain better speech recognition results, the Whisper model still requires manual specification of target languages during speech recognition.
The lower LID accuracy of the Whisper model can be attributed to the following factors:
\begin{itemize}
  \item
        Although the language is predicted using both the encoder and decoder networks, the language label is predicted as the first output of the decoder. Therefore, it can be considered that the system primarily relies on the acoustic feature and only implicitly uses the linguistic representation.
  \item
        By placing the language label at the beginning of the decoder network's inputs, it helps build a language-conditional decoder network that can easily process multilingual ASR tasks. However, this setting limits the potential for improving LID performance through the use of linguistic features. Especially, it makes fine-tuning the decoder for LID tasks difficult, because traditional decoder training often uses the true label as inputs.
  \item
        The Whisper model is designed mainly for multilingual speech recognition and translation, not specifically for LID tasks.
\end{itemize}
To enhance its LID performance, we first investigate a language embedding framework that explicitly extracts a richer linguistic representation. Subsequently, we also attempt to improve LID performance by further utilizing the implicit generation of linguistic representation.

\begin{figure*}
  \centering
  \includegraphics[width=480pt]{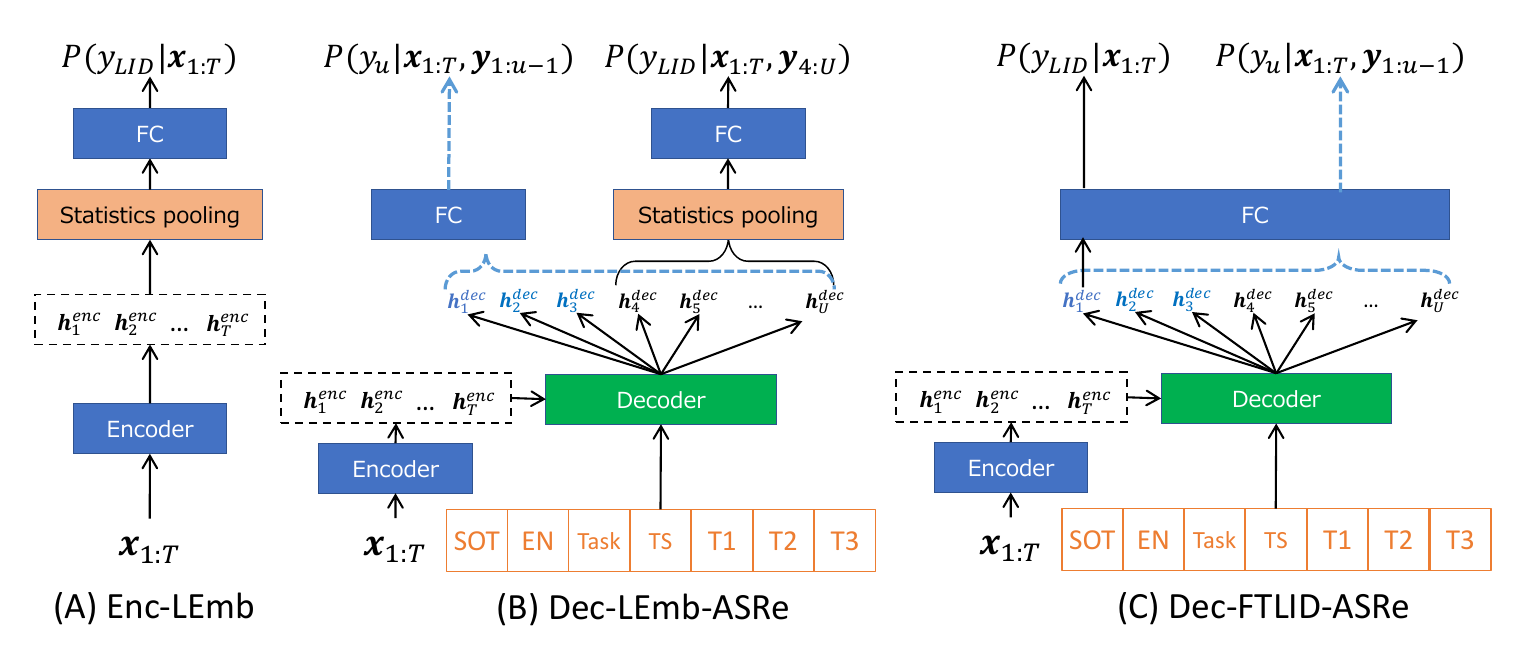}\\ \relax
  \caption{The proposed LID systems. (FC refers to the fully-connected layer, SOT is the Start of Transcription, EN represents English, Task stands for Transcription or Translation, TS indicates TimeStamp, T1,T2,T3 are tokens related to text.)}
  \label{fig.propose_method1}
\end{figure*}

\subsection{Language embedding-based LID}
The language embedding method are similar to the x-vector model \cite{snyer2018xvector4LID}, which comprises three core components: a frame-level feature extractor, a statistical pooling layer, and layers for utterance-level representation. In this study, the decoder outputs $\mathbf{h_{1:U}}$ (for encoder it is $\mathbf{h_{1:T}}$) serve as the acoustic representation. A statistics pooling layer transforms them into a fixed-dimension vector $\mathbf{h}_{p}$ through concatenation of the mean $\bm{\mu}$ and standard deviation $\bm{\sigma}$. The final stages utilize fully-connected hidden layers to manage the utterance-level representation and employ a softmax layer as output, with each output node representing a distinct language ID.
The operations can be described as follows:
\begin{equation}\label{eq:xvector.01}
  \begin{split}
    \mathbf{h}_{p} = Pooling(\mathbf{h}_{1:U}),
  \end{split}
\end{equation}
\begin{equation}\label{eq:xvector.02}
  \begin{split}
    \mathbf{h}_{fc} = FC_{2}(FC_1((\mathbf{h}_{p}))),
  \end{split}
\end{equation}
\begin{equation}\label{eq:xvector.03}
  \begin{split}
    P(y_{LID}|\mathbf{x}_{1:T}) = softmax(\mathbf{h}_{fc}).
  \end{split}
\end{equation}
Typically, a back-end classifier model such as the logistic regression (LR) classifier is used for classification with the output, before activations, from the first fully-connected layer that follows the statistics pooling layer. Before this process, dimension reduction is applied based on linear discriminant analysis (LDA) along with length normalization. In this work, we streamline the system by directly using the network output as the final prediction results, except in the case of the extended time delay neural network
(ETDNN) baseline system.

\subsubsection{Acoustic-based language embedding}
The encoder within the Whisper model essentially serves as a mechanism for acoustic feature extraction. As illustrated in Fig \ref{fig.propose_method1}.(A), we use the Encoder outputs as the acoustic features for Language Embedding (Enc-LEmb). The encoder output is fed into statistical pooling layers to extract utterance-level representations, which are then passed through fully-connected layers to predict language labels.
\subsubsection{Generative linguistic-based embedding}
An alternative approach involves utilizing decoder-based representation. The decoder network acts as a language model, thereby embedding linguistic information anticipated to contribute towards improving LID tasks. To explicitly extract this linguistic data, we encourage the decoder to undergo several iterations, yielding $N$ times outputs to gather ample linguistic information. In this study, we use the $N$ hidden outputs preceding the projection layer as the inputs for statistical language pooling. We remove the leading three output tokens as they are not linguistically related. We name this method Dec-LEmb.

\subsubsection{ASR-enhanced linguisitc-based embedding}
The previous approach aims to use linguistic information. However, since training optimizations focus only on language labels, the generative network could lose linguistic features. To overcome this, we propose a joint training framework that optimizes both LID and ASR tasks (Dec-LEmb-ASRe).
The optimization function is defined as follows:
\begin{equation}\label{eq:asr.lemb.03}
  \begin{split}
    L_{\text{Dec-LEmb-ASRe}} = (1-\lambda)L_{LID} + \lambda L_{ASR}
  \end{split}
\end{equation}
Where $\lambda$ is a weighting parameter that balances the losses of LID and ASR tasks.

During the training of the decoder network, the Whisper model uses the true language label as input tokens for the decoder, to build a language-conditional decoder. However, this setting allows the model to directly use this label information, making the LID task training process ineffective. To effectively train the decoder and avoid the issue of language label exposure, we propose two strategies:

\textbf{Fix-to-fix setting (EN2EN):} We fix the language ID input to the decoder and optimize the network to predict a fixed language ID. With this setting, the decoder will lose its language conditional characteristics. In this work, we consistently set the language ID to English, i.e., $EN$. The labels and inputs are as follows:
\begin{equation}\label{eq:asr.lemb.label01}
  \begin{split}
    \{EN, Task, TS, \dots\} \leftarrow \{SOT, EN, Task, TS, \dots\}
  \end{split}
\end{equation}
where 'Task' can refer to either 'Transcription' or 'Translation', while 'TS' stands for timestamp information. In our work, we didn't utilize any time alignment information.

\textbf{Fix-to-ground truth setting (EN2GT):} In this setup, we also use English as the input language. However, during prediction, we prompt the network to predict the true language label corresponding to the input speech. With these settings, we expect that the training will continue to use the encoder for LID, just as in the original Whisper model. The labels and inputs are described as follows:
\begin{equation}\label{eq:asr.lemb.label02}
  \begin{split}
    \{GT, Task, TS, \dots\} \leftarrow \{SOT, EN, Task, TS, \dots\}
  \end{split}
\end{equation}
where $GT$ means the ground truth language label.

\subsection{Implicit linguistic representation}

As we have discussed, the Whisper model is not specifically designed for LID tasks. To enhance LID performance, we evaluate two different learning methods - fine-tuning for LID and ASR-enhanced LID fine-tuning.

\subsubsection{LID fine-tuning}
Given that the original Whisper model includes an output for language label prediction, we can directly fine-tune this output to enhance the LID performance. This implies that we optimize $p(y_{LID}|\mathbf{x}_{1:T})$. Although this is the first output and seemingly only uses the encoder outputs, it does pass through the decoder. Thus, the output representation also encodes linguistic information, albeit in an implicit manner.

\subsubsection{ASR-enhanced LID fine-tuning}

Similar to the previous section, the fine-tuning process can lead to a deterioration of the linguistic representation that the ASR task has acquired. To mitigate this issue, we propose an ASR-enhanced learning method to optimize the network for improving LID tasks (Dec-FTLID-ASRe).
The target loss is defined as follows:
\begin{equation}\label{eq:strategy2.03}
  \begin{split}
    L_{\text{Dec-FTLID-ASRe}} = (1-w) * L_{LID} + w * L_{ASR}
  \end{split}
\end{equation}
In this equation, $L_{LID}$ is the cross-entropy loss based on language prediction output, corresponding to the first output label, i.e., $p(y_{LID}|\mathbf{x}_{1:T})$ in Fig. \ref{fig.propose_method1}.(C). $L_{ASR}$ is the original loss of the Whisper model, corresponding to $p(y_{u}|\mathbf{x}_{1:T},\mathbf{y}_{1:u-1})$. $w$ is the weight coefficient utilized to balance the two losses.
With this setup, we expect that the decoder network will preserve its linguistic representation during the optimization of the LID task.

\section{Experiments and results}
\subsection{Dataset and evaluation metrics}
To evaluate the proposed method, we utilized the large-scale multilingual LibriSpeech (MLS) dataset \cite{Pratap2020MLSAL}. The MLS dataset encompasses eight languages: English, German, Dutch, French, Spanish, Italian, Portuguese, and Polish.
The original MLS includes 44.5K hours of English, with an additional 6K hours divided among the seven other languages. To mitigate the impact of imbalanced training data across different languages, we gathered a maximum of 50,000 utterances for each language.
We constructed a validation set by randomly selecting 5,000 utterances and reserved the remaining data for the training set. For the test dataset, we prepared shorter utterance test data by segmenting fixed-length utterances of 1.0, 2.0, and 3.0 seconds from the original full-length test set.

In order to evaluate cross-domain data performance, we prepared two test datasets, one based on VoxLingua107 \cite{valk2021vox107slt} and the other on CommonVoice V6.1, each covering our target eight languages.
For the VoxLingua107 subset (Vox107), since the official evaluation dataset doesn't encompass data from all the target eight languages, we curated a new test dataset derived from the original training data. This test dataset was assembled by randomly selecting 20 videos per language, culminating in a final test dataset comprised of 5,437 utterances in total. The range of utterance numbers for each language spans from 344 to 1,259.
The CommonVoice subset (CV) was based on the its test dataset, with a maximum of 2,000 utterances selected per language for the test data.
Similar to the MLS dataset, shorter duration datasets of 1.0, 2.0, and 3.0 seconds were also prepared for the two out-of-domain datasets. Identification accuracy (Acc) is employed as the evaluation metric.

\subsection{Experimental setup}
\vspace{-0.5em}
We built several baseline systems, One baseline system is the ETDNN-based langauge embedding method, which employs an ETDNN for frame-level feature extraction \cite{snyder2019etdnn}. The two fully-connected layers after the pooling layer, each consisting of 512 neurons. We utilized 30-dimensional Mel Frequency Cepstral Coefficients (MFCC30) as features. To enhance the model's robustness, the SpecAugment technique was applied during training \cite{Park2019SpecAugment}. For classification, we used a logistic regression classifier after applying LDA-based dimension reduction and length normalization. The dimension of the LDA was set to 7.

Another baseline is a transducer-based system, we implemented the late statistic pooling-based fusion method by referring to \cite{shen2022transducerbased}. We used the Conformer (large) network for the Recurrent Neural Network Transducer (RNN-T) encoder and one Long Short-Term Memory (LSTM) layer for the prediction network, as outlined in \cite{conformer2020}. The Conformer encoder consists of 17 Conformer blocks, while the prediction network is made up of 640 LSTM memory cells. The outputs of the encoder and the prediction network were separately input into the statistic pooling and then merged for language identification. The statistical pooling layer and fully-connected layers are the same as those in the ETDNN system. For detailed setting, plese refer to \cite{shen2022transducerbased}. Same to the proposed method, we used the network output as the final recognition results.

In our proposed method, we carried out investigations utilizing both the Whisper base and large-v2 models \cite{whisper2022}. Both these models are based on transformer networks. The base model comprises 74M parameters, featuring 6 transformer blocks in both the encoder and decoder, each with a hidden dimension of 512. The large-v2 model contains 1550M parameters and includes 32x2 transformer blocks with a hidden dimension of 1280, making it a significantly large model for speech processing tasks. The input features used are 80-channel log-magnitude Mel spectrograms computed on 25ms windows with a stride of 10ms. The language embedding block is same to the ETDNN baseline. We opted not to use the SpecAugment technique in the proposed method.

We trained the model using the Adam algorithm with a warm restart learning rate scheduler during training \cite{Loshchilov2016Warmup}. For ETDNN, the learning rate was $0.001$, the mini-batch size was 512, with 21 training epochs. For the transducer-based method, the mini-batch size was 64 with an initial learning rate of $0.0001$, and a minimum learning rate of 1e-8.
For the proposed method, we tested initial learning rates of 1e-4, 1e-5, 1e-6, and 1e-7, adjusting them with a cosine annealing-based learning rate schedule, and finally used 1e-5 and le-6 for Dec-LEmb and Dec-FTLID methods, respectively. We set the number of training epochs to 10. Both the newly added parameters and the original parameters were optimized.

\vspace{-0.5em}
\subsection{Investigation and experimental results}
\vspace{-0.5em}

\begin{table*}[tb]
  \centering
  \caption{Experimental results (Acc \%) of baseline systems and proposed method on in-domain MLS test datasets, out-of-domain Vox107 and CV datasets.All the proposed methods are based on Whisper base model.}
  \setlength{\tabcolsep}{0.6em}
  \begin{tabular}{|l|c|c|c||c|c|c||c|c|c|c|c|c|c|c|c|c|} \hline
                                         & \multicolumn{3}{c||}{MLS} & \multicolumn{3}{c||}{Vox107} & \multicolumn{3}{c|}{CV}                                                                                                 \\ \hline
    Methods                              & 1s                        & 2s                           & 3s                      & 1s            & 2s            & 3s            & 1s            & 2s            & 3s            \\ \hline \hline
    ETDNN-MFCC30 (LEmb+LR)               & 76.8                      & 88.6                         & 92.9                    & 59.7          & 73.6          & 78.8          & 56.7          & 72.6          & 77.1          \\ \hline
    RNN-T AM (LEmb+pred)                 & 85.3                      & 94.4                         & 96.5                    & 66.8          & 77.2          & 80.7          & 68.5          & 82.8          & 85.9          \\ \hline
    RNN-T AMLM (LEmb+pred)               & \textbf{86.4}             & \textbf{94.9}                & \textbf{97.2}           & \textbf{68.9} & \textbf{81.1} & 84.8          & \textbf{69.8} & \textbf{84.8} & \textbf{87.7} \\ \hline
    Whisper (base)                       & 47.5                      & 83.4                         & 92.9                    & 49.5          & 78.5          & \textbf{87.6} & 18.3          & 64.4          & 83.2          \\ \hline\hline

    Enc-LEmb                             & 72.5                      & 89.3                         & 94.9                    & 59.8          & 79.8          & 87.1          & 57.6          & 78.7          & 85.1          \\ \hline
    Dec-LEmb                             & 74.3                      & 90.2                         & 95.4                    & 62.3          & 81.7          & 88.7          & 59.8          & 80.9          & 86.8          \\ \hline
    Dec-LEmb-ASRe-EN2EN ($\lambda$=0.3)  & 73.4                      & 91.6                         & 96.6                    & 57.1          & 78.0          & 85.4          & 55.6          & 78.0          & 84.9          \\ \hline
    Dec-LEmb-ASRe-EN2EN ($\lambda$=0.1)  & 75.4                      & 93.3                         & 97.2                    & \textbf{64.5} & \textbf{84.6} & \textbf{90.9} & 60.1          & \textbf{84.3} & \textbf{89.8} \\ \hline
    Dec-LEmb-ASRe-EN2EN ($\lambda$=0.05) & 75.0                      & 93.3                         & 97.5                    & 62.2          & 81.7          & 88.7          & 59.5          & 81.6          & 87.7          \\ \hline
    Dec-LEmb-ASRe-EN2GT ($\lambda$=0.1)  & \textbf{79.3}             & \textbf{93.8}                & \textbf{97.7}           & 62.8          & 81.0          & 87.3          & \textbf{61.3} & 82.0          & 87.7          \\ \hline\hline

    Dec-FT (Original)                    & 71.5                      & 82.4                         & 66.3                    & 59.7          & 80.0          & 83.7          & 56.9          & 77.7          & 74.2          \\ \hline
    Dec-FTLID                            & 79.5                      & 93.7                         & 97.7                    & 66.9          & 86.6          & 92.5          & 65.4          & 86.5          & 91.6          \\ \hline
    Dec-FTLID-ASRe (w=0.3)               & 79.7                      & 94.2                         & 97.9                    & 68.7          & 87.1          & 92.7          & 67.3          & 87.6          & 92.2          \\ \hline
    Dec-FTLID-ASRe (w=0.1)               & 80.4                      & 94.4                         & \textbf{98.0}           & 68.4          & 87.3          & 92.2          & 66.9          & 87.2          & 92.1          \\ \hline
    Dec-FTLID-ASRe (w=0.01)              & \textbf{81.8}             & \textbf{95.0}                & \textbf{98.0}           & \textbf{70.8} & \textbf{88.2} & \textbf{93.1} & \textbf{69.8} & \textbf{88.0} & 92.3          \\ \hline
    Dec-FTLID-ASRe (w=0.001)             & 81.7                      & 94.9                         & \textbf{98.0}           & 68.8          & 87.3          & 92.3          & 69.5          & \textbf{88.0} & \textbf{92.5} \\ \hline
  \end{tabular}
  \vspace{-3mm}
  \label{result.base}
\end{table*}

\begin{table*}[tb]
  \centering
  \caption{Experimental results (Acc \%) of the proposed method based on Whisper large-v2 model on in-domain MLS test datasets, out-of-domain Vox107 and CV datasets.}
  \setlength{\tabcolsep}{0.6em}
  \begin{tabular}{|l|c|c|c||c|c|c||c|c|c|c|c|c|c|c|c|c|} \hline
                                        & \multicolumn{3}{c||}{MLS} & \multicolumn{3}{c||}{Vox107} & \multicolumn{3}{c|}{CV}                                                                                                 \\ \hline
    Methods                             & 1s                        & 2s                           & 3s                      & 1s            & 2s            & 3s            & 1s            & 2s            & 3s            \\ \hline \hline
    Whisper (large-v2)                  & 61.7                      & 93.5                         & 98.0                    & 68.0          & 90.3          & 95.4          & 23.5          & 79.9          & 93.1          \\ \hline\hline
    Enc-LEmb                            & 79.0                      & 95.8                         & 98.3                    & 60.9          & 89.1          & 93.7          & 64.2          & 88.2          & 94.0          \\ \hline
    Dec-LEmb-ASRe-EN2EN ($\lambda$=0.1) & 87.5                      & 97.7                         & \textbf{99.2}           & 79.3          & \textbf{93.6} & \textbf{96.2} & 77.1          & 94.4          & \textbf{97.1} \\ \hline
    Dec-LEmb-ASRe-EN2GT ($\lambda$=0.1) & \textbf{87.7}             & 97.5                         & 99.0                    & \textbf{79.8} & 93.5          & 95.8          & \textbf{78.2} & \textbf{94.5} & \textbf{97.1} \\ \hline
    Dec-FTLID-ASRe (w=0.01)             & 82.7                      & 97.8                         & \textbf{99.2}           & 57.3          & 89.8          & 94.9          & 65.4          & 93.9          & 96.9          \\ \hline
    Dec-FTLID-ASRe (w=0.001)            & 86.1                      & \textbf{97.9}                & 99.1                    & 70.0          & 93.0          & \textbf{96.2} & 71.2          & 94.3          & 97.0          \\ \hline
  \end{tabular}
  \vspace{-3mm}
  \label{result.large}
\end{table*}

Table \ref{result.base} presents both the baseline results and the results of the proposed method applied to the three datasets.
For the RNN-T-based baseline systems, we evaluated configurations of both the acoustic feature (i.e., RNNT-AM) and a fusion of the acoustic and linguistic features (i.e., RNNT-AMLM).
We also evaluated the LID performance with the original Whisper base model.
From these results, it is evident that by incorporating linguistic representation, the RNNT-AMLM model achieved superior results compared to the RNNT-AM model, which solely uses the acoustic feature. The Whisper base model also exhibited comparable performance on relatively long utterances, particularly on the Vox107 datasets. However, its performance rapidly deteriorates with shorter utterances.
Next, we evaluated the language embedding-based approach using Whisper encoder-based features, where the encoder output was used for statistical pooling in language identification. From the results, we observed an improvement in performance compared to the original Whisper base model.
This improvement arises as the model's optimization focuses solely on the eight target languages, simplifying recognition compared to the original model designed to identify across 96 languages.

The two proposed methods (Dec-LEmb and Dec-FTLID) outperformed both the base Whisper model and the encoder-based language embedding method (Enc-LEmb). Even in comparison with the RNNT-AMLM model, our proposed methods outperformed it, with the exception of the 1-second in-domain dataset. This exception can be attributed to the fact that the RNNT-AMLM model was trained on short utterances, averaging 2-seconds in length. In the following, we will provide a detailed analysis of the proposed methods.
\vspace{-0.5em}
\subsubsection{Performance of the language embedding approaches}
For the Dec-LEmb and ASR-enhanced Dec-LEmb (Dec-LEmb-ASRe) methods, there are several parameters to consider, including the input-tokens to the encoder network, the output linguistic feature number $N$, and the weight hyperparameter $\lambda$.
Since the decoder operates as a language conditional model, any changes in the language input setting could significantly impact the extraction of linguistic features.
Thus, we evaluated several settings by modifying the language-label setting with both fix-to-fix (EN2EN) and fix-to-ground truth (EN2GT) configurations.
For task label, we fixed it to transcription.
To balance the LID loss and the original ASR-based loss from Whisper, we evaluated $\lambda$ in Eq. \ref{eq:asr.lemb.03} with the values of 0.05, 0.1, and 0.3. For the $N$ parameter, we fixed it at 10 for simplicity.

From the results, we observe that using the decoder-based feature, the LID performance was improved compared to both the base Whisper model and the encoder-based method. Moreover, by utilizing the ASR-enhanced training, the performance could be further improved, with the best results obtained when $\lambda$ was set to 0.1. Regarding the input-tokens setting, we found no significant difference between the two methods on in-domain data. However, for the out-of-domain data, the fix-to-fix configuration appeared to perform better. This may be due to the fact that the 'fix-to-fix' setting is easier to optimize, leading to a more stable optimization process.
\vspace{-0.5em}
\subsubsection{Performance of implicit linguistic approaches}
For the fine-tuning-based implicit linguistic approaches, we evaluated the weight parameter $w$ in Eq. \ref{eq:strategy2.03} with values of 0.001, 0.01, 0.1, and 0.3. It is noteworthy that, when we set $w$ to 0.0, the system optimizes the LID outputs and when set to 1.0, it optimizes the original ASR-based loss.

From the results, we found that simply using the original Whisper's optimization does not improve LID accuracy, i.e., results of Dec-FT(Original). This finding is mainly because the original Whisper model is primarily designed to enhance performance in ASR tasks. By focusing solely on optimizing the LID label, performance notably improved, achieving results comparable to the Dec-LEmb-ASRe method.
By further utilizing ASR optimization as a constraint (i.e., the Dec-FTLID-ASRe method), the performance was further improved. As the results show, when we assign a relatively small weight to ASR optimization (small $w$ value), the performance of LID improves significantly.

\subsubsection{Influence of model parameters}
Based on the investigation on the base model, we also evaluated the proposed methods using the large-v2 Whisper model. Given that we found a smaller learning rate to be more effective with the large-v2 model, we used a learning rate of 1e-6 for the Dec-LEmb-ASRe methods and 1e-7 for the Dec-FTLID-ASRe methods by refering to the performance with the base model. Other hyperparameters are based on the investigation of the base model.

Table \ref{result.large} presents the LID results for the Whisper large-v2 model and the proposed methods. The results indicate that methods based on the large-v2 model significantly outperform those based on the base model.
Unlike the base model, we discovered that the Dec-LEmb-ASRe method managed to narrow the gap with the Dec-FTLID-ASRe method, thereby achieving competitive results by using the large-v2 model.
As was the case with the results for the base model, the two language label input settings in Dec-LMeb-ASRe also achieved almost the same performance.
By comparing the results with the base model, we believe that the high prediction accuracy of the large-v2 provides sufficient information, allowing our Dec-LEmb-ASRe method to achieve greater improvements.

\section{Conclusions and Discussions}
In this study, we investigated methods to enhance LID tasks by leveraging the pre-trained Whisper generative model. We first analyze the limitations of the large-scale pre-trained Whisper model for LID tasks, and then propose a language embedding-based method and a fine-tuning based approach. To mitigate the forgetting problem during downstream tasks, we further use ASR-enhanced optimization to improve LID performance. We evaluated our method using both the base and large-v2 models. Our experiments demonstrated the effectiveness of the proposed method on both the in-domain and out-of-domain datasets.

From our investigations and experimental results, we observed that although the decoder network is a language conditional language model network, removing the language conditional assumption can also help to improve the performance of language embedding-based LID systems. Despite this, when designing downstream tasks, it's a good strategy to maintain the original settings as much as possible. This is particularly evident when using relatively smaller models.

Recent developments in large models have become a hot research topic. To understand the influence of large models on LID tasks, we compared the performance between base and large-v2 models.
We empirically assumed that methods based on explicit linguistic features would outperform those using implicit linguistic features.
However, this was not confirmed with the base model. With the large-v2 model, we found that the explicit linguistic-based method narrowed the gap and achieved competitively robust results.
This suggests, in line with trends observed in GPT-3 series models \cite{gpt3}, that an increase in model parameters significantly enhances accuracy. This improved accuracy aids the model in extracting high-quality linguistic features, thus bolstering the performance of the proposed language embedding methods.
We also observed distinct optimization settings between the large-v2 and base models, particularly in terms of learning rate and overfitting control.
For future work, a deeper understanding of these differences is necessary, particularly when dealing with relatively large models.

\section{Acknowledgements}
This work is supported by JSPS KAKENHI No.21K17776.

\bibliographystyle{IEEEbib}
\bibliography{lidbib}

\end{document}